# Spiking Inception Module for Multi-layer Unsupervised Spiking Neural Networks

**Mingyuan Meng**
The School of Electronics and Information Technology
Sun Yat-sen University
Guangzhou 510006, China
mengmy3@mail.sysu.edu.cn

**Xingyu Yang**
The School of Electronics and Information Technology
Sun Yat-sen University
Guangzhou 510006, China
yangxy266@mail2.sysu.edu.cn

**Shanlin Xiao**[*]
The School of Electronics and Information Technology
Sun Yat-sen University
Guangzhou 510006, China
xiaoshlin@mail.sysu.edu.cn

**Zhiyi Yu**[*]
The School of Electronics and Information Technology
Sun Yat-sen University
Guangzhou 510006, China
yuzhiyi@mail.sysu.edu.cn

*Abstract*—Spiking Neural Network (SNN), as a brain-inspired approach, is attracting attention due to its potential to produce ultra-high-energy-efficient hardware. Competitive learning based on Spike-Timing-Dependent Plasticity (STDP) is a popular method to train an unsupervised SNN. However, previous unsupervised SNNs trained through this method are limited to a shallow network with only one learnable layer and cannot achieve satisfactory results when compared with multi-layer SNNs. In this paper, we eased this limitation by: 1)We proposed a Spiking Inception (Sp-Inception) module, inspired by the Inception module in the Artificial Neural Network (ANN) literature. This module is trained through STDP-based competitive learning and outperforms the baseline modules on learning capability, learning efficiency, and robustness. 2)We proposed a Pooling-Reshape-Activate (PRA) layer to make the Sp-Inception module stackable. 3)We stacked multiple Sp-Inception modules to construct multi-layer SNNs. Our algorithm outperforms the baseline algorithms on the hand-written digit classification task, and reaches state-of-the-art results on the MNIST dataset among the existing unsupervised SNNs.

## I. INTRODUCTION

In recent years, artificial intelligence, especially neural network, has made great progress in machine perception and pattern recognition, reaching or even surpassing human in some application scenarios. Artificial Neural Network (ANN) has shown good performance in pattern recognition with deep learning, but ANN is highly computing-intensive. Therefore, many scholars begun to focus on the research of brain-inspired Spiking Neural Network (SNN) that is more biologically realistic and requires less computations [1]. The renewal of neural network is coming: neural network is evolving from the second generation, ANN, to the third generation, SNN [2].

In contrast to the traditional ANN whose information is represented by numerical values, the SNN uses spike trains to represent information. Although there is a gap existing between the performance of ANN and SNN on the cognition tasks, SNN's power consumption and execution latency are greatly reduced as a consequence of its data-driven, event-based style of computing [17]. Existing SNN algorithms can be classified into three types: *supervised* [4-5], *unsupervised* [7-12] and *conversion* [17-18]. *Supervised/Unsupervised* means the SNN is trained with/without using label information, while *conversion* denotes the algorithms of training an ANN first and then converting it into a SNN to circumvent the difficulties in training SNN directly. Currently, supervised/conversion algorithms achieve superior performance. Despite all this, in this paper we focus on the unsupervised SNN algorithms, because labeled data is expensive in many application scenarios, and unsupervised algorithms are considered to be more biologically plausible.

Competitive learning based on Spike-Timing-Dependent Plasticity (STDP) is a popular learning method to train an unsupervised SNN. However, the previous unsupervised SNNs trained through this method are limited to a shallow network with only one learnable layer [7-12] and can't achieve satisfactory results when compared with multi-layer SNNs. As is shown in Fig.1(a), baseline-FC module [7] and baseline-LC module [8] are two examples of SNN using STDP-based competitive learning. They are both 2-layer networks where the input layer is connected to the output layer in a Fully-Connected (FC) or Locally-Connected (LC) fashion, and the output neurons (i.e. the neurons in the output layer) compete with each other. The baseline modules are limited to 2-layer networks due to its low learning efficiency and low spiking intensity (The reasons for this limitation is thoroughly discussed in Section VI-B). To overcome this limitation and get a multi-layer unsupervised SNN, our main contributions in this paper are: 1)Inspired by the Inception module [20] in the ANN literature, we adopted the Split-and-Merge strategy (detailed in Section II-A) to propose a Spiking Inception (Sp-Inception) module (see Fig.1 (b)/(c)), which is trained through STDP-based competitive learning and outperforms the baseline modules on learning capability, learning efficiency and robustness. 2)We proposed a Pooling-Reshape-Activate (PRA) layer to reduce the output dimension and enhance the spiking intensity, thus making the Sp-Inception module stackable. 3) With the help of the PRA layer, we stacked multiple Sp-Inception modules to build a multi-layer SNN. Our multi-layer unsupervised SNN surpasses the baseline SNNs on the hand-written digit classification task, and achieves state-of-

This work was partly supported by National Key R&D Program of China under Grant 2017YFA0206200, Grant 2018YFB2202600 and National Nature Science Foundation of China (NSFC) under Grant 61674173, Grant 61834005, and Grant 61902443. * Shanlin Xiao and Zhiyi Yu both are corresponding authors of this paper.

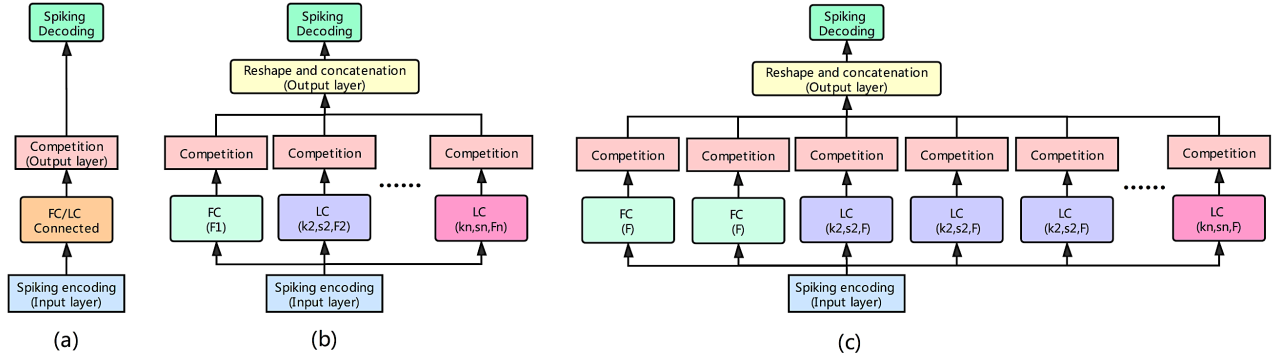

Fig. 1. SNN modules with STDP-based competitive learning. (a) Baseline-FC/LC module. (b) Sp-Inception module, naive version. (c) Sp-Inception module, balanced version.

the-art results on the MNIST dataset when compared with the existing unsupervised SNNs.

## II. RELATED WORK

### A. Inception Module

Inception module was first proposed by [20] and then evolved to many variants [21-22] in the ANN literature. The family of Inception module has demonstrated that carefully designed topologies are able to achieve compelling results with low theoretical complexity. An important common property of Inception modules is a Split-and-Merge strategy: The input is split into a few individual pathways with a set of specialized filters (e.g. $3 \times 3$, $5 \times 5$, $7 \times 7$ convolutional kernels), and then all pathways merge by concatenation. Under this strategy, the Inception modules can integrate multi-scale spatial information and reduce computing complexity. Besides, multiple Inception modules can be stacked together to form a very deep Convolutional Neural Network (CNN) [20].

Based on our review, this is the first paper to incorporate the principle of Inception module into multi-layer unsupervised SNNs. Our previous work [10] utilized an Inception-like multi-pathway unsupervised SNN, but this network is not stackable and limited to 3 layers. Xing et al. [18] built a spiking Inception architecture but trained it through a conversion approach.

### B. Unsupervised Learning and Multi-layer SNNs

Diehl et al. [7] (baseline-FC module) is one of the earliest papers to use STDP-based competitive learning to train an unsupervised SNN. It achieved the state-of-the-art result on the hand-written digits classification task in 2015 through a simple 3-layer FC SNN. Saunders et al. [8] (baseline-LC module) proposed to use LC connections to replace the FC connections in [7] and got a LC SNN with higher learning efficiency and robustness. Panda et al. [9] tried to incorporate the 'ability to forget' into [7] and proposed an Adaptive Synaptic Plasticity (ASP). She et al. [11] proposed to use a stochastic STDP in [7]. Our previous work [10] designed a fast-learning and high-robustness unsupervised SNN with a highly parallel Inception-like network architecture. Among them, Diehl et al. [7] and Saunders et al. [8] are chosen to be the baseline modules, because they can be regarded as the basis of our Sp-Inception module, all of us focusing on network topologies.

Since the SNN algorithms mentioned above are all limited to 2/3-layer networks with only one learnable layer, they are eventually exceeded by some multi-layer SNNs trained through supervised/conversion methods. Therefore, some scholars tried to discard competitive learning and relied on the unsupervised multi-layer convolutional SNNs [13-14]. These methods reach superior performance, but their SNNs are only used to extract image features and they require extra supervised classifier (e.g. SVM) to finish final classifications, which makes them regarded as semi-supervised rather than purely unsupervised algorithms in some literatures [8,10]. From our review, this is the first paper to use STDP-based competitive learning to train a multi-layer unsupervised SNN and get great performance improvements.

## III. BACKGROUND

In this section, we introduce the SNN background including computing unit model, spike coding scheme, and competitive learning theory. The methods described below are utilized widely [7-10] and are the default choices for the baseline modules and the Sp-Inception module in the experiments.

### A. Neuron and Synapse Model

Spiking neuron model is used to describe the behaviors of SNN's basic computing unit. Leaky Integrate-and-Fire (LIF) model is one of the most popular spiking neuron models and is used in this paper due to its simplicity. Moreover, there are many more complicated neuron models like Hodgkin-Huxley (HH) model [23] and Izhikevich model [24]. We only used a simple LIF model to build our SNN modules for emphasizing the effectiveness of our contributions. Following [7-8], the baseline-FC module uses a conductance-based LIF model, while the baseline-LC module, as well as our SP-Inception module, use a current-based LIF model. In this paper, the dynamics of the current-based LIF model is given by:

$$\tau_v \frac{dv(t)}{dt} = v_{rest} - v(t) + RI(t) \quad (1)$$

$$\tau_I \frac{dI(t)}{dt} = -I(t) + \sum_i^N G(s_i) w_i F(n_i, t) \tau_I \quad (2)$$

In (1), $v(t)$ is the voltage (membrane potential), $v_{rest}$ is the neuron's resting voltage, $I(t)$ denotes the total input current to the neuron, $R$ is synaptic resistance, and $\tau_{v/I}$ is the time constant of $v(t)/I(t)$. In (2), $N$ is the number of presynaptic neurons connected to the neuron, $s_i$ is the synapse between the neuron and presynaptic neuron $n_i$, and $w_i$ is the synaptic weight of

$s_i$. The function $G(s_i)$ is equal to 1/-1 when $s_i$ is excitatory/inhibitory synapse. The function $F(n_i, t)$ is equal to 1 when $n_i$ fires a spike at time $t$, otherwise it's equal to 0. According to (1) and (2), the $v(t)$ and $I(t)$ decay exponentially to $v_{rest}$ and 0 respectively when no presynaptic neuron fires spikes. At the occurrence of a spike from an excitatory /inhibitory synapse, the $I(t)$ increases/decreases by the weight of this synapse, thus leading to the change of membrane potential $v(t)$. When the $v(t)$ reaches or exceeds a threshold $\theta$, the neuron fires a spike to downstream neurons and the $v(t)$ is reset to a voltage $v_{reset}$. After firing a spike, the neuron does not integrate input spikes for a refractory period $T_{ref}$. Moreover, the adaptive repolarization scheme proposed in [10] is adopted to accelerate the network's learning. A homoeostasis mechanism in [7] is used to ensure that no neuron can emit excessive spikes and dominate the firing activity. The homoeostasis is an adaptive threshold scheme as follows:

$$\tau_\theta \frac{d\theta(t)}{dt} = v_{thres} - \theta(t) + \tau_\theta \theta_{plus} F(n, t) \qquad (3)$$

where $n$ denotes the neuron itself and $\tau_\theta$ is time constant of $\theta(t)$. Concretely, each time the neuron $n$ fires a spike, the threshold $\theta$ increases by a constant $\theta_{plus}$, or it decays exponentially to a voltage $v_{thres}$.

We also used a synapse model to describe the dynamics of synaptic weight. STDP model was widely used in unsupervised SNNs [7-12] and evolved to many variants (e.g. additive STDP [3], triplet STDP [15]). In this paper, following [8], we used a very basic STDP rule as follows:

$$\Delta w = \begin{cases} \eta_{post} x_{pre} & \text{when postsynaptic spike} \\ -\eta_{pre} x_{post} & \text{when presynaptic spike} \end{cases} \qquad (4)$$

where $\eta_{pre}/\eta_{post}$ are the pre/postsynaptic learning rates, and $x_{pre}/x_{post}$ are pre/postsynaptic traces. The update of synaptic weight $w$ occurs when pre/postsynaptic neuron fires a spike (named pre/postsynaptic spike). The synaptic traces are used to record the timing of the previous spikes. The $x_{pre}/x_{post}$ is reset to 1 when the pre/postsynaptic spike is fired, or they decay exponentially to 0 with $\tau_{pre}/\tau_{post}$ as their time constant. Note that we set $\eta_{post} \gg \eta_{pre}$ to emphasize the effects of pre-synaptic neurons on post-synaptic neurons.

### B. Input Encoding and Spike Decoding

Information in SNN is represented by discrete spike trains, but each pixel value in the input image is an analog value. Therefore, we need to conduct input encoding to convert the analog pixel values into the discrete spike trains. In this paper, we adopted a popular rate-based encoding scheme used in [7-10]. The input neuron (i.e. the neuron in the input layer) is a generator of Poisson-distributed spike trains, and each input neuron corresponds to a pixel of the input image. Each pixel value is encoded into a Poisson-distributed spike train whose average rate is equal to the pixel value multiplied by an encoding parameter $\lambda$. Moreover, we adopted the adaptive encoding scheme used in [7], because the module might be insensitive to some images and larger $\lambda$ is required sometimes.

Since the output of SNN module is also spike trains, we need to decode the output into recognizable inference results when applying trained SNN into image classification tasks. In general, the unsupervised SNNs with rate-based input encoding use vote-based methods for spike decoding [7-10]. Following [7-8], the baseline-FC module uses a classic vote-based decoding scheme, while the baseline-LC module utilizes a n-gram method [19] (In detail, n = 2). In Sp-Inception module, we adopted the VFA decoding layer proposed in [10], because we found that this method outperforms other vote-based decoding schemes when implemented in our Sp-Inception module.

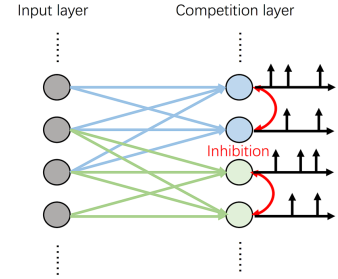

Fig. 2. An illustration of competitive learning: The neurons in the competition layer are interconnected by fixed inhibitory synapse and compete with other neurons sharing the same RF (i.e. having the same set of presynaptic neurons).

### C. STDP-based Competitive Learning

The STDP rule described in Section III-A adjusts the synaptic weight based on the relative timing of pre/postsynaptic spikes. Under the STDP model, if a presynaptic spike tends to occur immediately before a postsynaptic spike, the synaptic weight is made bigger; If a presynaptic spike tends to occur immediately after a postsynaptic spike, the synaptic weight is made smaller. Based on the former phenomenon, STDP-based competitive learning was proposed and widely used in many unsupervised SNNs [7-12]. That's also the reason why we set $\eta_{post} \gg \eta_{pre}$ to emphasize the former phenomenon.

The principle of STDP-based competitive learning is each neuron learns and represents a prototype which is initialized randomly and gradually becomes similar to the real input through learning. Every time a training sample is input, the neurons compete with each other and only the one whose represented prototype is more similar to this input sample can fire spikes and increase its STDP-modifiable (i.e. modified based on STDP rule) synaptic weights to make its represented prototype even more similar to this input sample. Those winner neurons are finally used to predict the class of the testing sample because they respond so much to the samples similar to their represented prototypes. The baseline-FC/LC modules and our Sp-Inception module all are based on this learning principle. As is shown in Fig. 2, each neuron in the competition layer is connected to the neurons sharing the same Receptive Field (RF) (i.e. sharing the same set of presynaptic neurons) with fixed inhibitory synapses, which means that each neuron competes for learning to represent a prototype appearing in real input samples.

## IV. METHOD

In this section we illustrate our main contributions: in Section IV-A we detail the design of Sp-Inception module which can be used as an image classifier directly; in Section IV-B we show the method of stacking multiple Sp-Inception modules to build a multi-layer SNN.

## A. Spiking Inception (Sp-Inception) Module

Inspired by the Inception module in the ANN literature [19-22], we designed a naive version of Sp-Inception module. As is shown in Fig. 1(b), the naive version is composed of several processing pathways connected to the input layer in a FC/LC fashion. Each pathway computes and performs competition independently. Then, they are reshaped to one-dimension and concatenated together. The input neurons are the generators of Poisson-distributed spike trains, and the neurons in the competition layer are LIF models. Following the principle of competitive learning described in Section III-C, the input layer and competition layers are connected with the excitatory STDP-modified synapses, and the neurons in the competition layer are interconnected with the fixed inhibitory synapses. The LC connection has the same topology with the convolutional connection but without using shared weight. We use a kernel size $k$ and a stride $s$ to define RF (only square kernel is used in this paper), and use a $F$ to denote feature map number. For FC connection, we also use the $F$ to denote neuron number, because the FC connection actually is a special case of the LC connection where the kernel size is equal to the input layer size. In this case, a neuron in the FC connections can be regarded as a feature map.

In the naive version of Sp-Inception module, we make sure there is at least one FC pathway to cope with global feature. Then, other topology settings including pathway number, FC/LC choice, and parameters $k/s/F$ are decided empirically. However, we found that there is a problem on the naive version: the learning speed of each pathway is unbalanced and the overall learning speed is hindered by its slowest pathway. Especially when a pathway's $F$ is set to be much larger than ones of other pathways, this pathway would learn much more slowly, but sometimes this situation is inevitable for achieving better module performance. Since the learning efficiency of a pathway mainly depends on its feature map number $F$ (shown in Section V-A(2)), to solve this problem we refined the naive version by: 1)We divided a pathway with large $F$ into several pathways with smaller $F$. E.g. A pathway with $F = 1600$ can be divided into four pathways with $F = 400$. By doing so, the learning speed of this pathway improves with negligible performance degradation. 2)We restricted each pathway's $F$ to be same. By doing so, the learning speed of each pathway becomes balanced. Finally we got a balanced version of Sp-Inception module in Fig. 1(c).

## B. Multi-module Architecture

To overcome the limitations of previous unsupervised SNNs with only one learnable layer, we explored the possibility of building a multi-layer SNN by stacking Sp-Inception modules. However, although Inception module has shown some amazing improvements (shown in Section V-A), there are still two bottlenecks existing in the process of stacking modules: 1) The Sp-Inception module's output dimension is much more than its input dimension, which means that the data dimension will explode if we directly stack the Sp-Inception modules without dimension reduction; 2)The Sp-Inception module's low output spiking intensity makes it unpractical to stimulate its downstream module (detailed in Section VI-B).

In order to solve these two bottlenecks, we proposed a Pooling-Reshape-Activate (PRA) layer to connect two Sp-Inception modules. Fig. 3 shows an example on how to stack Sp-

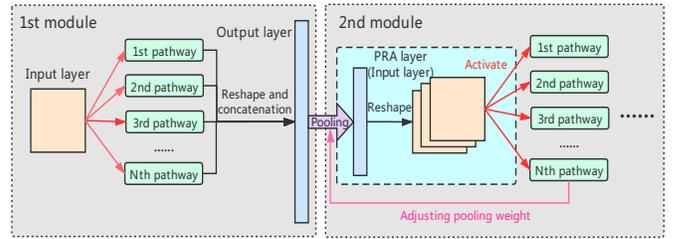

Fig. 3. An illustration of how to stack two Sp-Inception modules through a PRA layer. The PRA layer is inserted between two Sp-Inception modules and performs Pooling-Reshape-Activate.

Inception modules through a PRA layer. As is shown in Fig. 3, the PRA layer is inserted between two modules and also is used to be the input layer of latter module (in place of original input layer). What the PRA layer does are: 1) Reduce the output dimension of former module by pooling with size $P$; 2) Reshape one-dimensional layer into three dimensions $28 \times 28 \times C$ ($C$ is the input channels of latter module); 3) Activate the latter module through an adaptive pooling mechanism. The PRA layer is composed of LIF models, so its spiking intensity can increase if its pooling connections' weights are enlarged. In adaptive encoding scheme used in first module, the encoding parameter $\lambda$ is adjusted according to the module's output spiking intensity. Similarly, in the adaptive pooling mechanism, the pooling connections of the PRA layer share the same weigh $w_p$ and this $w_p$ is adjusted according to the module's output spiking intensity. When an image is input, if any module's output spiking intensity is too low, this module's $w_p$ (or $\lambda$ if this module is the first one) will increase by a constant and then this image is input again. With the help of PRA layer, we can make sure that each module's output spikes are enough for activating its downstream modules.

Theoretically, this architecture can be expanded to a very deep multi-layer SNN if computing resources are sufficient. But we also need to be careful to avoid overfitting. In the following experiments, we stacked four Sp-Inception modules and got good results on it. More details about this four-module architecture are shown in Section V-B.

## V. EXPERIMENTS

In this section, we evaluate the proposed Sp-Inception module (a single module directly used as a classifier) and the multi-layer SNN consisting of multiple stacked Sp-Inception modules. All training and testing procedures are implemented with the MNIST dataset [16]. In the MNIST dataset, there are 70,000 hand-written digital images labeled from 0 to 9 and each image is $28 \times 28$ in size. Among them, 60,000 images are in the training set and 10,000 images are in the testing set. More experimental details including hyperparameter settings, training procedures, experimental environment, and the implementations of baseline modules are presented in the Appendix A.

### A. Sp-Inception Module Evaluation

#### 1) Learning Capability

In Table I, we report the testing results of SNN modules on the MNIST and the number of neurons/synapses used by them (denote by $n_{neuron}/n_{synapse}$). Here we use $(k, s) \times F$ to denote kernel size, stride, and feature map number of a LC topology and use only $F$ to denote neuron number of a FC topology (As is mentioned in Section IV-A, each neuron in FC topology can be regarded as a feature map). As is shown in Table I, the modules with larger $F$ exhibit better testing results but utilize more neurons and synapses. From this perspective, The Sp-Inception module is more efficient on the resource usage. Compared with the baseline modules, Sp-Inception modules can achieve higher testing results with less neurons and synapses used (e.g. Sp-Inception I-III). If we allow the Sp-Inception module to utilize more resources (equal to or even more than ones used by the baseline modules), they can get significantly improved testing results (e.g. Sp-Inception IV-VI). Note that we didn't test the Sp-Inception module with $F > 448$, because the Sp-Inception modules with $F \leq 448$ have exhibited satisfactory results, and keeping increasing $F$ will make the Sp-Inception module utilize excessive resources but gain slight improvement.

*2) Learning Efficiency*

To evaluate the learning efficiency of Sp-Inception module, in Fig. 4 we report the testing results of the modules trained with varying number of training iterations. In Fig. 4(a), we compare the modules with similar learning capability: baseline-FC IV, baseline-LC III, and Sp-Inception III all exhibit close learning capabilities of about 94.8% results when fully trained. Then, we compare the modules with the same feature map number $F$ in Fig. 4(b): baseline-FC I, baseline-LC II, and Sp-Inception V all have 400 feature maps; baseline-FC II and baseline-LC III both have 800 feature maps. Since Sp-Inception module with $F = 800$ is impractical for implementation, it's not tested in Fig. 4(b).

Through Fig. 4, we find that: 1) The learning efficiency of a module mainly depends on its feature map number $F$. E.g. In Fig.

TABLE I. COMPARISON OF LEARNING CAPABILITY AMONG THE BASELINE MODULES AND SP-INCEPTION MODULE

| Module | Id | Topology | $n_{neuron}$ | $n_{synapse}$ | Result |
|---|---|---|---|---|---|
| Baseline-FC | I | FC, $F = 400$ | 400 | 473K | 87.80% |
| | II | FC, $F = 800$ | 800 | 1267K | 90.12% |
| | III | FC, $F = 1600$ | 1600 | 3814K | 91.94% |
| | IV | FC, $F = 6400$ | 6400 | 45977K | 94.88% |
| Baseline-LC | I | LC, $(16,6) \times 100$ | 900 | 320K | 91.36% |
| | II | LC, $(16,6) \times 400$ | 3600 | 2361K | 93.97% |
| | III | LC, $(16,6) \times 800$ | 7200 | 7603K | 94.83% |
| | IV | LC, $(16,6) \times 1000$ | 9000 | 11304K | 95.02% |
| Sp-Inception | I | $\begin{pmatrix} FC, F = 112 \\ LC, (24,4) \times 112 \\ LC, (16,6) \times 112 \end{pmatrix}$ | 1568 | 778K | 93.36% |
| | II | $\begin{pmatrix} FC, F = 224 \\ LC, (24,4) \times 224 \\ LC, (16,6) \times 224 \end{pmatrix}$ | 3136 | 1909K | 94.59% |
| | III | $\begin{pmatrix} (FC, F = 300) \times 4 \\ (LC, (24,4) \times 300) \times 2 \\ LC, (16,6) \times 300 \end{pmatrix}$ | 6300 | 4904K | 94.86% |
| | IV | $\begin{pmatrix} FC, F = 448 \\ LC, (24,4) \times 448 \\ LC, (16,6) \times 448 \end{pmatrix}$ | 6272 | 5224K | 95.57% |
| | V | $\begin{pmatrix} (FC, F = 400) \times 4 \\ (LC, (24,4) \times 400) \times 2 \\ LC, (16,6) \times 400 \end{pmatrix}$ | 8400 | 7379K | 95.62% |
| | VI | $\begin{pmatrix} FC, F = 448 \\ LC, (24,4) \times 448 \\ LC, (16,6) \times 448 \\ LC, (10,6) \times 448 \end{pmatrix}$ | 13440 | 9153K | 95.85% |

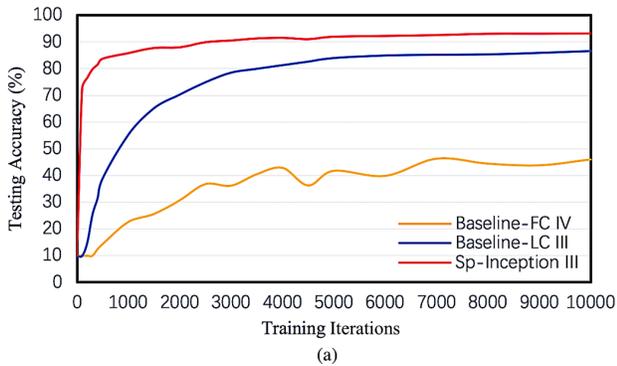
(a)

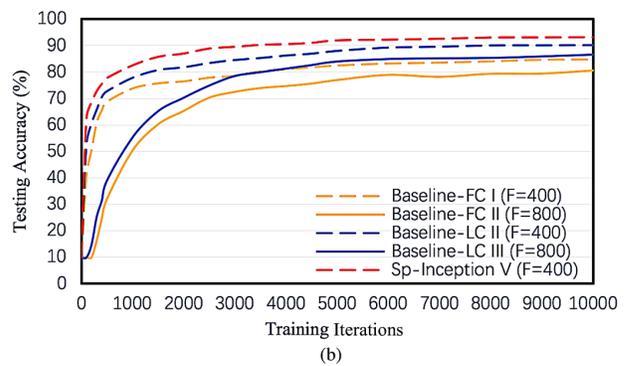
(b)

Fig. 4. Testing results of the modules trained with varying number of training iterations. (a) Comparison among the modules with similar learning capability. (b) Comparison among the modules with the same feature map number $F$.

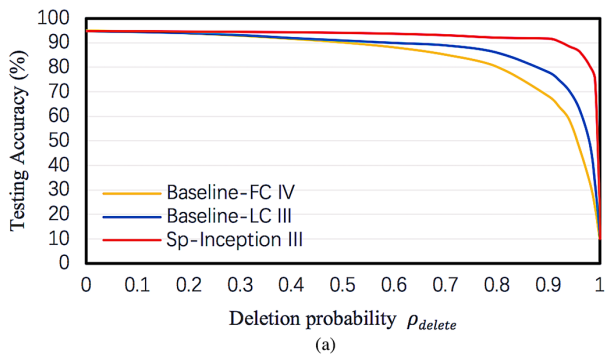
(a)

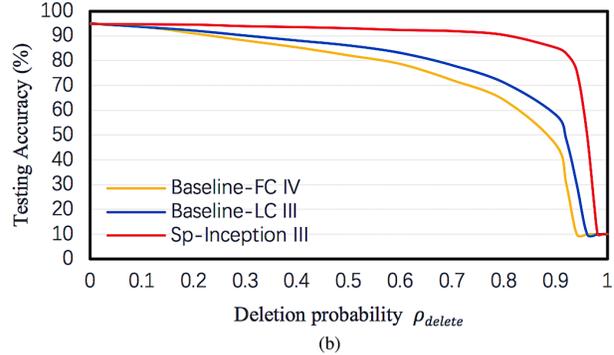
(b)

Fig. 5. Testing results of the modules whose neurons/synapses are randomly deleted with varying probability $\rho_{delete}$. (a) Neuron deletion. (b) Synapse deletion.

4(b), baseline-FC I, baseline-LC II, and Sp-Inception V show similar learning efficiency, and baseline-FC II, baseline-LC III also show close learning efficiency, even though their final learning capabilities are totally different. 2) Compared with the baseline modules, Sp-Inception modules achieve better testing results with less feature map number needed, thus making Sp-Inception III ($F = 300$) learn much faster than baseline-FC IV ($F = 6400$) and baseline-LC III ($F = 800$). In Fig. 4(a), when modules are trained with 10000 iterations, the baseline-FC/LC only achieve about 45%/86% testing results, while the Sp-Inception III has achieved more than 93% testing result. The first find above also explains why we finally use the balanced version of Sp-Inception module (Fig. 1(c)) instead of the original naive version (Fig. 1(b)).

*3) Robustness*

Robustness is network's resistance against external damage and interferences. To evaluate it, we followed the evaluation scheme used in [8,10]. Concretely, we randomly deleted some learnable synapses or output neurons of the trained baseline-FC IV/baseline-LC III/Sp-Inception III with probability $\rho_{delete}$, then we report their testing results in Fig. 5. As can be seen in Fig. 5, the three modules without any neuron/synapse deletion reach the results of about 94.8% on the MNIST, while increasing $\rho_{delete}$ leads to a smooth degradation in testing results.

Fig. 5(a) shows the testing results after deleting neurons with probability $\rho_{delete}$ and our Sp-Inception module exhibits the highest robustness. Sp-Inception III maintains nearly 90% and 80% results respectively with even 92% and 98% of neurons deleted, while baseline-FC IV and baseline-LC III maintain nearly 80% and 85% results with only 80% of neurons deleted. Similarly, Fig. 5(b) shows the testing results after deleting synapses with probability $\rho_{delete}$ and our Sp-Inception module still exhibits the highest robustness. Sp-Inception III maintains nearly 90% and 80% results respectively with even 80% and 92% of synapses deleted, while baseline-FC IV and baseline-LC III only maintain nearly 70% and 65% results with only 80% of synapses deleted. This experiments demonstrate that Sp-Inception module has higher robustness against external damage and destruction. It can work well even through most of its learnable synapses or computing neurons break down.

*B. Multi-module Archiecture Evaluation*

Due to the limitations of experimental time and computing resources, we only stacked four Sp-Inception modules and trained them as a whole multi-layer SNN. In the four-module architecture, we can use the output of any middle modules to perform inference and see the improvements got by stacking each module. In Table II, we report the testing results along with the number of neurons/synapses we've used after we stack each module. $P$ denotes the pooling size of PRA layer. As is shown in Table II, the testing result increases as more modules are stacked, which shows our multi-module architecture works.

We also compare our method with the existing unsupervised SNN algorithms on the MNIST dataset. The testing results of compared algorithms are found in the corresponding references and then directly listed in Table III. Note that the semi-supervised algorithms such as [13-14] are excluded because the comparison becomes unfair when they use extra supervised classifier like SVM. As is shown in Table III, our methods reach the state-of-the-art results on the MNIST dataset. Admittedly, our method's result is not the best one in Table III. But our contributions are merely about architecture design, and we only used very basic neuron model, spiking coding, and learning rule. Therefore, our method can be in conjunction with other more advanced SNN components (e.g. ASP [9], stochastic STDP [11]) to build a better recognition system.

TABLE II. TESTING RESULTS OF A FOUR-MODULE ARCHITECTURE

| Network | First/stacked Module | $n_{neuron}$ | $n_{synapse}$ | Result |
|---|---|---|---|---|
| 1st module | Sp-Inception I | 1568 | 778K | 93.36% |
| +2nd module | + Sp-Inception II, $P = 1$ | 4707 | 3894K | 95.17% |
| +3rd module | + Sp-Inception IV, $P = 2$ | 10976 | 11533K | 96.03% |
| +4th module | + Sp-Inception VI, $P = 2$ | 24416 | 23818K | 96.48% |

TABLE III. COMPARISON OF TESTING RESULTS ON THE MNIST DATASET AMONG THE EXISTING UNSUPERVISED SNN ALGORITHMS

| Paper | Description | Result |
|---|---|---|
| Diehl et al. 2015 [7] | 3-layer FC SNN (Baseline-FC module) | 95.00% |
| Saunders et al. 2019 [8] | 2-layer LC SNN (Baseline-LC module) | 95.07% |
| Panda et al. 2017 [9] | 3-layer FC SNN with Adaptive Synaptic Plasticity (ASP) | 96.80% |
| Meng et al. 2019 [10] | Inception-like multi-pathway 3-layer SNN | 95.64% |
| She et al. 2019 [11] | 3-layer FC SNN with stochastic STDP | 96.10% |
| Lammie et al. 2018 [12] | 2-layer FC SNN (FPGA neuromorphic system) | 94.00% |
| Ours | Single Sp-Inception module | **95.85%** |
| Ours | Multi-layer SNN consisting of four Sp-Inception modules | **96.48%** |

## VI. DISCUSSION

*A. Why does the Sp-Inception module outperform the baseline modules?*

The Sp-Inception module, as well as the Inception module [20-22], have some common properties with visual cortex of human brain. For instance, in the same area of visual cortex, neurons possess different RF sizes and work efficiently in parallel [6]. It's well known that visual cortex performs amazing learning capability, learning efficiency, and robustness, which motivates us to answer this question from this perspective.

In the Sp-Inception module, multiple processing pathways with different topologies (e.g. FC fashion, LC fashion with different RF sizes) enable the module to collect multi-scale spatial information, thus helping improve its learning capability. Moreover, this design enables each processing pathway to learn and compute in parallel, which significantly enhances the learning efficiency and robustness. Actually, Saunders et al. [8] has proven that the baseline-LC outperforms the baseline-FC on learning efficiency and robustness. We took a further step to combine baseline-LC module and multiple processing pathways.

*B. Why is the Sp-Inception module, rather than the baseline modules, stackable?*

Baseline-FC module is not stackable partly because it learns too slow. Referring to [7] and our experiments, the baseline-FC

TABLE IV. AVERAGE SPIKING INTENSITY OF THE BASELINE MODULES AND SP-INCEPTION MODULE

| Module | Baseline-FC | Baseline-LC | Sp-Inception | Sp-Inception +PRA layer |
|---|---|---|---|---|
| Output Intensity (spikes / iteration) | 7.84 | 58.26 | 142.83 | 2294.87 |
| Input Intensity (spikes / iteration) | 3451.26 | 3284.85 | 2936.65 | |

module with 6400 neurons needs to be trained with 900,000 iterations, which is ten times slower than Sp-Inception module. If the baseline-FC modules are stacked, learning efficiency is even lower. However, merely this answer is not enough because baseline-LC module learns much faster but is still unstackable. To fully answer this question, in Table IV we report the average output spiking intensity of baseline-FC IV, baseline-LC III, Sp-Inception IV, and the average input spiking intensity when they are trained. It's well shown that, compared with the input spiking intensity, the output spiking intensity of baseline module is too low to motivate another module. However, with the help of the PRA layer, the output spiking intensity of Sp-inception module is high enough to motivate its downstream modules.

### C. Why do the stacked Sp-Inception modules outperform the single Sp-Inception module?

To answer this question, we looked inside the network and saw what each stacked module receives when processing an image. In Fig. 6, we visualized the input spiking distribution of each stacked module in a four-module architecture. Fig. 6(a)/(b) show the average input spiking distributions for four stacked modules when processing the images of digit 0/1, and each dark spot denotes the possibility of module receiving input spikes in this position (the darker it's, the more likely the module receives input spikes). Fig. 6(c) shows the overlap of two input spiking distributions. As is shown in Fig. 6, we find that the overlap area becomes smaller with more modules stacked. This gives us a possible assumption: The input received by later module is more spatially distinctive because the input spikes corresponding to the different classes tend to occur in different positions, which makes it easier for later module to classify the input.

To further validate this assumption, we calculated Mean Spatial Distribution Similarity (MSDS, detailed in Appendix B) of the input spiking maps belonging to the different classes, and visualized them in Fig. 7. In each iteration, we recorded the input spike number in each position and used it as pixel value to draw an input spiking map. In Fig. 7, if the MSDS value is closer to 1, the input spiking maps have more similar spatial distribution and vice versa. We also report the average MSDS values alongside the figure. As is shown in Fig. 7, the MSDS of input spiking maps tends to decrease with more module stacked, which validates our above assumption.

### VII. CONCLUSION

In this paper, we proposed a Sp-Inception module to construct the unsupervised SNNs for image classification tasks. The Sp-Inception module not only can be used directly as an unsupervised classifier but also is stackable to build multi-layer unsupervised SNNs. In Section V, we evaluated the proposed Sp-Inception module and the multi-layer SNN composed of stacked Sp-Inception modules. It's shown in Section V that: 1)

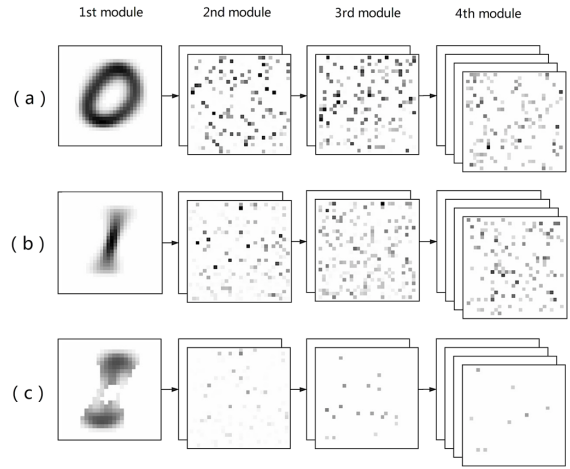

Fig. 6. Input spiking distributions of each stacked modules. (a) Input images of digit 0. (b) Input images of digit 1. (c) Overlap between (a) and (b).

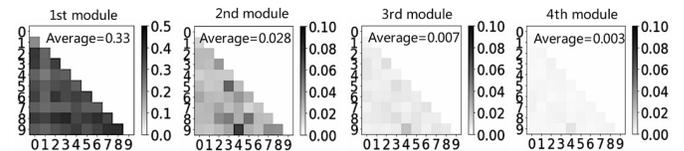

Fig. 7. MSDS matrixs over the input spiking maps of 10 classes from 0 to 9.

The Sp-Inception module can outperform the baseline modules on learning capability, learning efficiency, and robustness; 2) The multi-module architecture exhibits better classification performance than a single Sp-Inception module, and reaches the state-of-the-art results on the MNIST dataset when compared with the existing unsupervised SNN algorithms. In Section VI, we discussed the reasons why we got these improvements: 1) Multiple processing pathways with different kernels (i.e. RF sizes) and high parallelism of Sp-Inception module help explaining its superior performance; 2) The spiking intensity is tested to answer why the Sp-Inception module, rather than the baseline modules, is stackable; 3) The input of each module in a four-module architecture is visualized to show each module's effect on its downstream module.

The Sp-Inception module eases the limitations of previous shallow unsupervised SNNs (the baseline modules). Its high learning efficiency and the proposed PRA layer make the Sp-Inception module stackable to form a multi-layer unsupervised SNN. Note that the results in this paper are not the ceiling of our method. Better performance could be achieved if more Sp-Inception modules are stacked or better hyperparameters are found. Moreover, in order to emphasize the effectiveness of our contributions (merely about architecture design), other SNN components (e.g. neuron model, synapse model, spiking coding scheme, etc.) are very simple. Our contributions can be in conjunction with other more sophisticated SNN components to get stronger SNN algorithms.

### APPENDIX

#### A. Experimental Details

Our experiments ran on a 8-core system with 32GB RAM in an Ubuntu environment. All codes are based on an open-source simulator, Brian [25]. We strictly followed the parameter settings, network architectures, and training procedures used in [7-8] to implement the baseline modules except that: 1) The inhibitory layer in [7] was replaced by the inter-connections of excitatory layer; 2) The input size in [8] was changed from $20 \times 20$ to $28 \times 28$. For our Sp-Inception module, all hyperparameters are empirical values (listed in Table V). We estimated the scopes of parameter values according to the related references [7-12] and then decided them through the cross-validation in which we randomly chose 10,000 images from the training set as a validation set.

TABLE V. HYPERPARAMETER SETTINGS IN THE EXPERIMENTS

| Hyperparameter | Description | Value |
|---|---|---|
| $\eta_{post}$ | Postsynaptic learning rate | 0.01 |
| $\eta_{pre}$ | Presynaptic learning rate | 0.0001 |
| $\tau_{pre}/\tau_{post}$ | Time constant of $x_{pre}/x_{post}$ | 20ms |
| $v_{thres}$ | Threshold voltage | -52mv |
| $v_{rest}$ | Resting voltage | -65mv |
| $v_{reset}$ | Resetting voltage | -65mv |
| $\theta_{plus}$ | Increment for adaptive threshold | 0.05mv |
| $T_{ref}$ | Time length of refractory period | 5ms |
| $\tau_{v/I}$ | Time constant of $V(t)/I(t)$ | 100ms/1ms |
| $\tau_\theta$ | Time constant of adaptive $\theta$ | $10^7$ms |

After the hyperparameters were decided, we used all 60,000 images of training set to train our network. We adopted the training procedures used in [10] and the weight normalization used in [8]. Moreover, in order to speed up the simulation of multi-module architecture, we forwarded the spiking activities module by module rather than simulated the whole multi-module network simultaneously. Specifically, we recorded the neuron activities of the former module and then used them to stimulate the latter module. But this change won't cause any problem because each module computes locally and is not interfered by its downstream modules.

*B. Mean Spatial Distribution Similarity (MSDS)*

To measure the mean similarity of spatial distributions of the images belonging to two sets, we define a Mean Spatial Distribution Similarity (MSDS). Assume that there are $N_1$ images in set $C_1$ and $N_2$ images in set $C_2$. Then, each image is normalized so that its mean pixel value is equal to 1. The MSDS between $C_1$ and $C_2$ is defined as follows:

$$MSDS(C_1, C_2) = \frac{1}{N_1 \times N_2} \sum_i^{N_1} \sum_j^{N_2} SDS(C_1(i), C_2(j)) \quad (5)$$

$$SDS(I_1, I_2) = 1 - \frac{1}{\sum_k^S I_1(k) + \sum_k^S I_2(k)} \sum_k^S |I_1(k) - I_2(k)| \quad (6)$$

where SDS is the spatial distribution similarity between two images, $S$ is the size of image $I_1/I_2$, and $I_1(k)/I_2(k)$ denotes the $k^{th}$ pixel value of image $I_1/I_2$.